\documentclass[runningheads]{llncs}
\usepackage[T1]{fontenc}
\usepackage{graphicx}
\usepackage{listings}
\usepackage{algorithm}
\usepackage{algcompatible}
\usepackage{algpseudocode}

\usepackage{amsmath,amssymb}
\usepackage{cleveref}
\usepackage{stmaryrd}
\usepackage{booktabs}
\usepackage{tabularx}
\usepackage{dirtytalk}
\usepackage{multirow}
\usepackage{adjustbox}
\usepackage{bm}
\usepackage{xcolor,colortbl}
\usepackage{tikz}

\usepackage[acronym]{glossaries}
\newacronym{us}{US}{Ultrasound}
\newacronym{fps}{fps}{frames per second}
\newacronym{ef}{LVEF}{Left Ventricular Ejection Fraction}
\newacronym{es}{ES}{End-Systolic}
\newacronym{ed}{ED}{End-Diastolic}
\newacronym{sv}{SV}{Systolic Volume}
\newacronym{cdm}{CDM}{Cascaded Diffusion Model}
\newacronym{edm}{EDM}{Elucidated Diffusion Model}
\newacronym{ml}{ML}{Machine Learning}
\newacronym{mri}{MRI}{Magnetic Resonance Imaging}
\newacronym{ct}{CT}{Computed Tomography}
\newacronym{vae}{VAE}{Variational Auto-Encoder}
\newacronym[longplural={Denoising Diffusion Probabilistic Models},plural={DDPMs}]{ddpm}{DDPM}{Denoising Diffusion Probabilistic Model}
\newacronym[longplural={Denoising Diffusion Implicit Models},plural={DDIMs}]{ddim}{DDIM}{Denoising Diffusion Implicit Model}
\newacronym{adm}{ADM}{Ablated Diffusion Model}
\newacronym{pndm}{PNDM}{Pseudo Numerical methods for diffusion models}
\newacronym[longplural={Generative Adversarial Networks},plural={GANs}]{gan}{GAN}{Generative Adversarial Network}
\newacronym{vqvae}{VQ-VAE}{Vector Quantized Variational Autoencoder}
\newacronym{lidm}{LIDM}{Latent Image Diffusion Model}
\newacronym{lvdm}{LVDM}{Latent Video Diffusion Model}
\newacronym[longplural={Stochastic Differential Equations},plural={SDEs}]{sde}{SDE}{Stochastic Differential Equation}
\newacronym[longplural={Ordinary Differential Equations},plural={ODEs}]{ode}{ODE}{Ordinary Differential Equation}
\newacronym{vdm}{VDM}{Video Diffusion Model}
\newacronym{fid}{FID}{Fréchet Inception Distance}
\newacronym{is}{IS}{Inception Score}
\newacronym{fvd}{FVD}{Fréchet Video Distance}
\newacronym{mae}{MAE}{Mean Average Error}
\newacronym{lpips}{LPIPS}{Perceptual Similarity}
\newacronym{dyn}{Dyn}{EchoNet-Dynamic}
\newacronym{ped}{Ped}{EchoNet-Pediatric}
\newacronym{syn}{Syn}{EchoNet-Synthetic}
\newacronym{a4c}{A4C}{Apical 4 Chamber}
\newacronym{psax}{PSAX}{Parasternal Short Axis}
\newacronym[longplural={Latent Diffusion Models},plural={LDMs}]{ldm}{LDM}{Latent Diffusion Model}

\newcommand{\ua}{\uparrow}
\newcommand{\da}{\downarrow}

\makeatletter
\newcommand*{\inlineequation}[2][]{%
  \begingroup
    \refstepcounter{equation}%
    \ifx\\#1\\%
    \else
      \label{#1}%
    \fi
    \relpenalty=10000 %
    \binoppenalty=10000 %
    \ensuremath{%
      #2%
    }%
    ~\@eqnnum
  \endgroup
}
\makeatother

\begin{document}
%
%\title{EchoNet-Synthetic: A Diffusion-Generated Open-Source Dataset for Echocardiogram Ejection Fraction Regression}

% \title{Privacy-preserving Score-Based Synthesis for Safe and Efficient Medical Image Data Sharing}
\title{EchoNet-Synthetic: Privacy-preserving Video Generation for Safe Medical Data Sharing}

\titlerunning{EchoNet-Synthetic}
\author{
Hadrien~Reynaud\inst{1,2} \and 
Qingjie~Meng\inst{3} \and 
Mischa~Dombrowski\inst{4} \and 
Arijit~Ghosh\inst{4} \and 
Thomas~Day\inst{5,6} \and
Alberto~Gomez\inst{5,7} \and
Paul~Leeson\inst{7,8} \and 
Bernhard~Kainz\inst{2,4}
}
%index{Reynaud, Hadrien}
%index{Meng, Qingjie}
%index{Dombrowski, Mischa}
%index{Ghosh, Arijit}
%index{Gomez, Alberto}
%index{Leeson, Paul}
%index{Kainz, Bernhard}
%
\institute{
UKRI CDT in AI for Healthcare, Imperial College London, London, UK \and
Department of Computing, Imperial College London, London, UK \email{hadrien.reynaud19@imperial.ac.uk} \and
School of Computer Science, Univeristy of Birmingham, Birmingham, UK \and
Friedrich--Alexander University Erlangen--N\"urnberg, DE \and
School of BMEIS, King's College London, London, UK \and
Guy's and St Thomas' NHS Foundation Trust, London, UK \and
Ultromics Ltd, Oxford, UK \and 
John Radcliffe Hospital, Cardiovascular Clinical Research Facility, Oxford, UK
}
\authorrunning{H. Reynaud et al.}
\maketitle              % typeset the header of the contribution
\begin{abstract}
To make medical datasets accessible without sharing sensitive patient information, we introduce a novel end-to-end approach for 
generative de-identification of dynamic medical imaging data. 
Until now, generative methods have faced constraints in terms of fidelity, spatio-temporal coherence, and the length of generation, failing to capture the complete details of  dataset distributions.
We present a model designed to produce high-fidelity, long and complete data samples with near-real-time efficiency and explore our approach on a challenging task: generating  echocardiogram videos. 
We develop our generation method based on diffusion models and introduce a protocol for medical video dataset anonymization. 
As an exemplar, we present EchoNet-Synthetic, a fully synthetic, privacy-compliant echocardiogram dataset with paired  ejection fraction labels. 
As part of our de-identification protocol, we evaluate the quality of the generated dataset and propose to use clinical downstream tasks as a measurement on top of widely used but potentially biased image quality metrics.
Experimental outcomes demonstrate that EchoNet-Synthetic achieves comparable dataset fidelity to the actual dataset, effectively supporting the ejection fraction regression task.
Code, weights and dataset are available at \url{https://github.com/HReynaud/EchoNet-Synthetic}.

\end{abstract}

\section{Introduction}\label{sec:intro}

\noindent
Medical datasets play a crucial role for learning-based medical image analysis~\cite{ouyang2020video,rueckert2016learning} and form the basis for a potential future medical foundation model~\cite{bommasani2021opportunities}. 
However, the confidential nature of medical data, coupled with concerns over privacy and extensive individual patient-consent requirements, often restricts dataset releases. 
To address these challenges, we propose a new protocol for generative medical dataset de-identification.
In our protocol, we first train a generative model, \emph{e.g.}, diffusion models, to learn the data distribution from a real training set.
Then, we sample our generative model to generate a synthetic dataset, with the same properties as the real training set, \emph{e.g.}, diversity, quantity, population statistics, etc.
This synthetic dataset is then filtered to preserve patient privacy and to safeguard against model memorization.
To verify the quality of this privacy-preserving synthetic data, we train a downstream model (regression, classification, segmentation, ...) on the generated data, and evaluate it on a held out set of real data.
We compare the model's performance against the same downstream model trained on the real training data.
If their performance difference is within an acceptable range, it proves that the synthetic dataset is a valid substitute to the real training dataset, and can be shared as such without data privacy constraints.

\noindent\textbf{Contributions: }
(1) We introduce the first \gls{lvdm} capable of generating high-fidelity, long echocardiogram videos at near real-time speeds.
(2) We propose a comprehensive protocol for generating useful medical datasets. This protocol emphasizes the importance of dataset quality, validated through the training of downstream models (\emph{e.g.}, ejection fraction regression) and ensures that synthetic datasets can effectively support medical imaging research and clinical translation.
(3) We release EchoNet-Synthetic, a pioneering fully synthetic echocardiography dataset that maintains the quality and diversity required for effective downstream model training, while protecting patient privacy. Models trained on EchoNet-Synthetic exhibit performance metrics comparable to those trained on real datasets, thereby validating the efficacy of our dataset generation protocol.

\noindent\textbf{Related Works:}
Video generation is an important research area within computer vision.
Recently, several works started studying diffusion models for video generation 
\cite{an2023latent,blattmann2023align,harvey2022flexible,ho2022imagen,ho2022video,hoeppe2022diffusion,khachatryan2023text2video,luo2023videofusion,nikankin2022sinfusion,reynaud2023feature,singer2022make,voleti2022masked,wang2023videofactory,yang2022diffusion,zhou2022magicvideo}.
\phantom{-}These diffusion-based video generation methods can achieve outstanding modelling capabilities, but they suffer from excessive computational requirements.
% and their controllability is often limited to text prompt engineering. 
To make them more tractable, concurrent works explore \glspl{ldm} for image generation~\cite{rombach2022high} and for video generation~\cite{blattmann2023align,he2022latent,yu2023video,zhou2022magicvideo}. 
VideoLDM~\cite{blattmann2023align} extends \glspl{ldm} to high-resolution video generation by turning pre-trained image \glspl{ldm} into video generators by inserting temporal layers. 
Latent Video Diffusion Models (LVDM)~\cite{he2022latent} proposes a hierarchical diffusion model operating in the video latent space, allowing long video generation through a second model.

In the field of ultrasound generation, some works focus on physics-based simulators~\cite{jensen2004simulation,shams2008real} while others use \gls{gan}-based methods for individual images. 
These works condition their models on \gls{mri}, \gls{ct}~\cite{teng2020interactive,tomar2021content} or simulated~\cite{gilbert2021generating,tiago2022data} images. 
Other works focus on ultrasound video generation. 
\cite{liang2022sketch} presents a motion-transfer-based method for pelvic ultrasound videos, while \cite{reynaud2022d} introduces a causal model for echocardiogram video generation. 
\cite{reynaud2023feature} introduces a diffusion model-based video generation method for echocardiogram synthesis, which however requires extensive sampling times% and lacks temporal consistency.
In this work, we build upon \glspl{ldm} for fast and temporally-consistent echocardiogram generation.

\iffalse
\gls{ef} is an important clinical biomarker for the diagnosis of heart failure, and regression models can be leveraged for fast and systematic \gls{ef} estimation in echocardiograms. 
Compared to a variety of works, \emph{e.g.},~\cite{mokhtari2022echognn,reynaud2021ultrasound}, using an R2+1D model~\cite{tran2018closer} trained over fixed-length videos is still the most reproducible \gls{ef} regression method~\cite{ouyang2020video}. 
In contrast to the existing echocardiogram video generation methods, which only use generative metrics, we further evaluate our echocardiogram generation method on the downstream \gls{ef} regression task, ensuring the clinical translation of our method. 
\fi

Privacy issues have recently gained a lot of attention due to diffusion models' ability of to memorize training samples \cite{carlini2023extracting}.
Current approaches focus on two aspects, privatizing the generative model \cite{dombrowski2023quantifying} and filtering out the generated samples that raise privacy concerns \cite{dar2024unconditional,packh_auser2022deep}.
In this work, we focus on the second approach to remove samples that can be linked back to the training set.

\glsreset{lvdm}

\section{Method}\label{sec:method}
\begin{figure}[t]
    \centering
    \includegraphics[width=\textwidth]{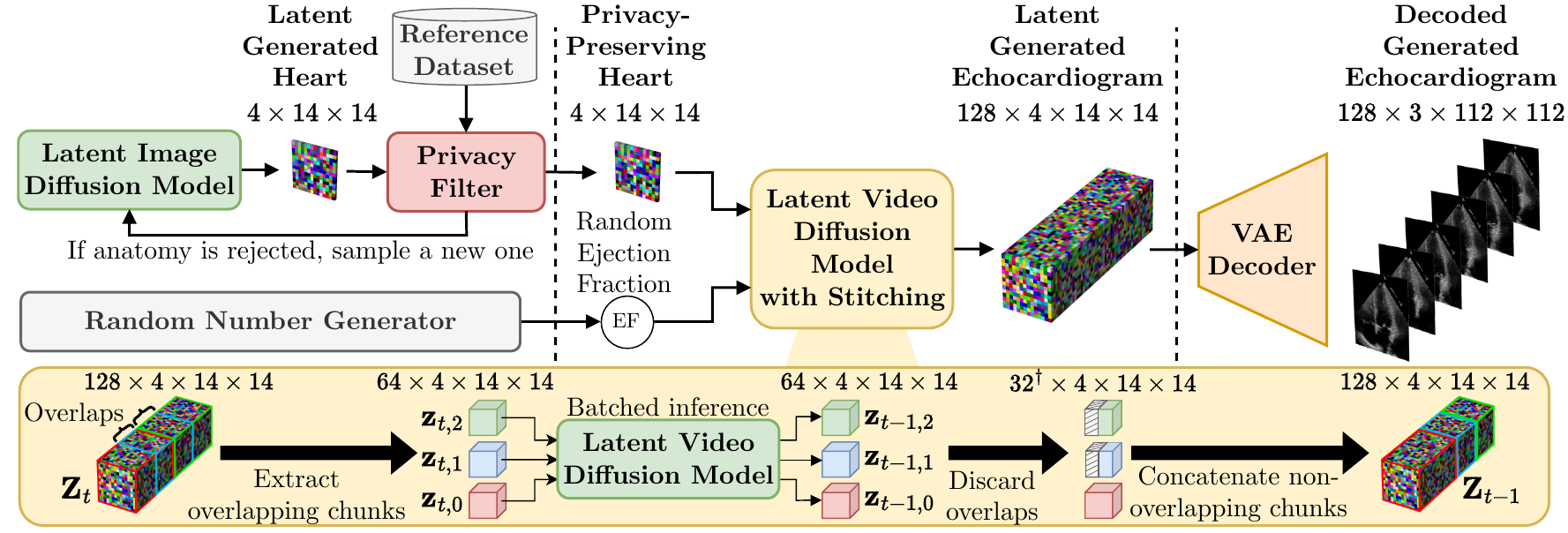}
    \caption{
    Our inference video generation pipeline, consisting of a latent image diffusion model with a privacy filter, a latent video diffusion model with video stitching and a \gls{vae} decoder.
    All dimensions are given in (time $\times$) channels $\times$ height $\times$ width. 
    $\dagger$ In the video stitching method (yellow block), the first half of $\mathbf{z}_{t-1,0}$ is not discarded.
    }
    \label{fig:generation}
\end{figure}

The goal of our method is to generate videos suitable to train downstream models while preserving patient privacy and allowing unlimited video duration.
In this section, we present our de-identification protocol, consisting of (1) video generation, (2) privacy filtering, (3) video stitching and (4) evaluation on a downstream task.
\Cref{fig:generation} summarizes our video generation pipeline. 

%fix vae acronym not working for some reason
\glsreset{vae}
\noindent\textbf{Video Generation:}
In this work, we implement an \gls{ldm} pipeline for video generation.
To do so, we train three models, a  \gls{vae}, a \gls{lidm} and a \gls{lvdm}, independently.

We start by training a \gls{vae} on an image reconstruction task.
The \gls{vae} is made of four downsampling blocks, based on attention and convolutional layers.
Given an input image \( I \in \mathbb{R}^{C \times H \times W} \), the variational autoencoder (VAE) encodes it to produce two 3D latent tensors: a mean \(\mu\) and a standard deviation \(\sigma\), such that \(\{\mu, \sigma\} \in \mathbb{R}^{4 \times H/8 \times W/8}\).
These latent tensors are used to define an n-dimensional Gaussian distribution, which is sampled to produce latent representations of the image.
From this latent representation, the \gls{vae} decoder reconstructs the input image.
The \gls{vae} is trained by using a combination of MSE, LPIPS~\cite{zhang2018unreasonable} and GAN~\cite{goodfellow2014generative} loss, and a KL-divergence loss which constrains the latent space to be Gaussian distributed.
After the \gls{vae} is trained, we encode the datasets into the \gls{vae} latent space, and train the diffusion models on these latent representations.

The \gls{lidm} and \gls{lvdm} are trained as discrete-time \glspl{ddpm}~\cite{ho2020denoising} with a v-prediction~\cite{salimans2022progressive} objective.
The noising process of the \gls{ddpm} is defined as a Markov chain, where
\begin{align}
p(\mathbf{z}_{1:T}|\mathbf{z}_0) = \prod_{t=1}^{T} p(\mathbf{z}_t|\mathbf{z}_{t-1}).
\end{align}
The noise at each step is Gaussian, such that $p(\mathbf{z}_t|\mathbf{z}_{t-1}) = \mathcal{N}(\mathbf{z}_t; \sqrt{1-\beta_t}\mathbf{z}_{t-1}, \beta_t\mathbf{I})$, and $\beta_t$ specifies the noise variance.
Here, $\mathbf{z}_t$ is the noisy latent and $T$ is the total number of diffusion steps.
For the denoising process, a neural network is trained to estimate the velocity $\mathbf{v}$, such that $\mathbf{v} = \alpha_t\epsilon - \sigma_t\mathbf{z}$, and updates $\mathbf{z}$ through
\begin{align}
\mathbf{z}_{t-1} = \alpha_t\mathbf{z}_t - \sigma_t\mathbf{v}_{\theta}(\mathbf{z}_t),\label{eq:diffusion_reverse}
\end{align}
with $\alpha_t$ and $\sigma_t$ as time-dependent coefficients, and $\mathbf{v}_{\theta}$ as the v-prediction model.

The difference between the \gls{lidm} and \gls{lvdm} resides in their backbone neural networks and their training strategy.
The \gls{lidm} is implemented as an unconditional UNet with residual downsampling blocks containing convolutional layers and self-attention layers. 
We set the number of residual blocks to four, with channel sizes 128, 256, 256, 512. The \gls{lidm} is trained to generate latent images by minimizing an MSE loss over its outputs and the \gls{vae} pre-encoded images. %with the \gls{vae} model, with an MSE loss.
The \gls{lvdm} consists of a Spatio-Temporal UNet~\cite{blattmann2023stable}, with four residual blocks and channel sizes 128, 256, 256, 512. The residual blocks contain convolutional layers and cross-attention layers, both with space-time separation. 
To train the \gls{lvdm} we use an MSE loss, and condition the model on an encoded real heart image and a given \gls{ef} score.

During inference, we combine the three models into a video generation pipeline, as shown in \Cref{fig:generation}.
We start by sampling Gaussian noise, and use the \gls{lidm} to progressively generate a random latent heart.
Then, this latent heart, a randomly sampled \gls{ef} score and some Gaussian noise are passed to the \gls{lvdm}.
The \gls{lvdm} produces a latent video of the latent heart, where the motion of the heart is conditioned by the given \gls{ef}.
Finally, we use the \gls{vae}-decoder to decode the latent video frames into a pixel-space echocardiogram.

\noindent\textbf{Video Stitching:}
Generating long videos with diffusion models remains a challenging task, even in a latent space, because of the training computational cost.
Naively extending the number of video frames during inference does not yield realistic temporal consistency. This is because the attention layers in the backbone model cannot handle more than the time window it was trained on. 
To circumvent this limitation and allow the generation of longer videos during inference, we split and denoise a long noisy latent video into overlapping pieces, and then stitch them back together (see \Cref{fig:generation}).
As these overlapping pieces are aware of the content of their predecessor, our method ensures seamless temporal continuity in long videos.
Formally, we start from a noisy latent video $\mathbf{Z}_{t} \in \mathcal{R}^{l_v \times c \times h \times w}$, where $l_v$ is the video length (\emph{e.g.}, 128).
The \gls{lvdm} $\mathbf{v}_{\theta}$ is trained on videos with length $l_m$ (\emph{e.g.}, 64), and $l_v > l_m$.
We define an overlapping factor $o = \frac{l_m}{2}$ and split the $\mathbf{Z}_{t}$ tensor into $k$ overlapping chunks, $k = \frac{l_v - l_m}{o}+1$. 
This produces $\mathbf{z}_{t,i} = \mathbf{Z}_t[i \cdot (l_m-o) : i \cdot (l_m-o)+l_m] , \forall i \in \{0, 1, \ldots, k-1\}$. 
Then, we denoise these overlapping noisy latent videos with \Cref{eq:diffusion_reverse}, \emph{i.e.}, $\mathbf{z}_{t-1,i} = \alpha_t \cdot \mathbf{z}_{t,i} - \sigma_t \cdot \mathbf{v}_{\theta}(\mathbf{z}_{t,i})$. 
Finally, to reconstruct $\mathbf{Z}_{t-1}$ from all $\mathbf{z}_{t-1,i}$, we discard the 
overlaps
over the time axis, \emph{i.e.}, for $\forall i>0$, $\mathbf{z}_{t-1,i} = \mathbf{z}_{t-1,i}[o:l_m] $, and then, we concatenate all updated $\mathbf{z}_{t-1,i}$ over the time axis.

\noindent\textbf{Privacy filtering:}
Since the \gls{lvdm} only animates the heart images it is given, ensuring that those images are privacy compliant is enough to guarantee that the corresponding generated videos are also privacy compliant.
Thus, we apply the privacy filter on the latents generated by the \gls{lidm}. %, rather than on the decoded images.
Specifically, we train a re-identification model~\cite{packh_auser2022deep} on the encoded real training set.
The re-identification model is trained with a contrastive loss.
We set the positive pairs as different frames from the same video and the negative pairs as two frames from different videos.
The penultimate layer of the re-identification model is used to compute the distance between two samples. 
Following \cite{dar2024unconditional}, we use the Pearson correlation score to measure that distance.
During inference, we use the first frame of each video to compute distances.
We calculate the distance between all real training samples and all real validation samples.
From this distribution of distances, we find the threshold $\tau$, where 95\% of the distances are above $\tau$.%\cite{dar2024unconditional}.
Then, we compute the distances between all synthetic and all real training samples.
For each synthetic sample, if the shortest distance to a real samples is under $\tau$, we consider the synthetic sample to be `memorized' and exclude it.

\noindent\textbf{Evaluation on downstream tasks: }
Current generated medical datasets are usually evaluated with standard generative metrics such as \gls{fid}, \gls{fvd} and \gls{is}. However, these metrics do not directly correlate with performance on downstream tasks such as classification, regression, or segmentation. 
To address this, following our de-identification protocol, we propose to generate synthetic datasets and assess them on established downstream benchmarks to confirm their efficacy. 
Specifically, we first use our privacy-preserving video generation pipeline to generate a synthetic echocardiogram dataset.
Then, we compare the performance of an \gls{ef} regression model trained on real data, against the same model trained on our generated dataset.
After training, both models are evaluated on the same real test set to ensure that the surrogate dataset's distribution aligns with the real data, thereby validating its quality and diversity.

\section{Experiments}\label{sec:experiments}
\noindent\textbf{Data: }
To facilitate reproducibility, we use the publicly available echocardiogram datasets, \gls{dyn} \cite{ouyang2020video} and the \gls{ped} \cite{reddy2023video}, as real datasets.
Every echocardiogram in the real datasets has a manually labelled \gls{ef}.
\gls{dyn} contains exclusively \gls{a4c} views, split into training, validation and test sets.
\gls{ped} contains \gls{a4c} and \gls{psax} views, with each view split into 10 folds.
To match the datasets structure, we split \gls{ped} into \gls{ped}(\gls{a4c}) and \gls{ped}(\gls{psax}), and arbitrarily use the first 8 folds for training, the $9^{\text{th}}$ for validation and the $10^{\text{th}}$ for test.
All videos have 3 channels, and we preprocess all videos to greyscale. 

\noindent\textbf{Model training: }
Our \gls{vae} is trained on all the frames from the training sets of \gls{dyn} and \gls{ped}.
We train the model from scratch for 5 days on $8 \times$A100~(80GB), with a total batch size of 256, and a learning rate of $5e-4$. 
The \gls{vae} compresses the frames from $3 \times 112 \times 112$ to $4\times14\times14$.

We train three independent \gls{lidm}s for \gls{dyn}, \gls{ped}(\gls{a4c}) and \gls{ped}(\gls{psax}), by using the \gls{vae}-encoded train sets of each dataset.
The backbone UNet operates in a $4\times16\times16$ configuration to allow three downsampling steps, so we use replication padding on the latents produced by the \gls{vae} to match those dimensions.
The three \gls{lidm}s are trained for 24h on a single A100~(80Gb), with a batch size of 256, and a learning rate of $3e-4$. 

We train a re-identification model for each real dataset on a single A100 with batch size 128 for 1000 epochs. All other parameters are taken from \cite{dar2024unconditional}.

The \gls{lvdm} is trained on all the training sets of the three datasets.
During training, we sample a pair of latent video and \gls{ef} from the training set.
We use a random frame from the latent video as the input heart.
Conditioned on this input heart and the paired \gls{ef}, our \gls{lvdm} is trained to reconstruct the sampled video with a fixed length of 64 frames.
We train the model for 2 days with a learning rate of $1e-4$ on $8 \times$A100 (80GB), with a total batch size of 128.

\noindent\textbf{Model evaluation: }
We evaluate the \gls{vae} by computing the usual reconstruction metrics and generative metrics. 
The results in \Cref{table:vae} show that our \gls{vae} is capable of reconstructing echocardiogram frames with high fidelity, and thus learns a meaningful latent space.  

% \begin{table}[t]
%     \centering
%     \caption{Performance of the \gls{vae}.
%     % Left: paired reconstruction metrics. Right: dataset-wise reconstruction metrics. 
%     The \gls{ped} datasets do not have enough videos with more than 128 frames to enable FVD$_{128}$ computation.}
%     \label{table:vae}
%     \resizebox{0.9\columnwidth}{!}{%
%     \begin{tabular}{lcccccccccc}
%         & \multicolumn{5}{c}{\textbf{Reconstruction Metrics}} && \multicolumn{4}{c}{\textbf{Generative Metrics}} \\
%          & {MSE}$\da$ & {MAE}$\da$ & {SSIM}$\ua$ & {PSNR}$\ua$ & {LPIPS}$\da$ && {FID}$\da$ & {FVD$_{16}$}$\da$ & {FVD$_{128}$}$\da$ & {IS}$_{\pm\mathbf{std}}$$\ua$ \\ 
%         \cmidrule{2-6}\cmidrule{8-11}
%         \gls{dyn}               & $3e-3$ & 0.03 & 0.78 & 24.9 & 0.08 && 16.9 & 87.4 & 69.2       & $2.33_{\pm{0.13}}$ \\
%         \gls{ped}(\gls{a4c})    & $1e-3$ & 0.02 & 0.86 & 28.9 & 0.07 &&  6.9 & 21.8 & -  & $2.90_{\pm{0.09}}$ \\
%         \gls{ped}(\gls{psax})   & $2e-3$ & 0.02 & 0.85 & 28.5 & 0.08 &&  8.1 & 22.1 & -  & $3.16_{\pm{0.13}}$    \\
%     \end{tabular}
%     }
% \end{table}

\begin{table}[t]
    \centering
    \caption{Performance of the \gls{vae}.
    % Left: paired reconstruction metrics. Right: dataset-wise reconstruction metrics. 
    The \gls{ped} datasets do not have enough videos with more than 128 frames to enable FVD$_{128}$ computation.}
    \label{table:vae}
    \resizebox{1.0\columnwidth}{!}{
    \begin{tabular}{lcccccccccc}
        & \multicolumn{5}{c}{\textbf{Reconstruction Metrics}} && \multicolumn{4}{c}{\textbf{Generative Metrics}} \\
         & {MSE}$\da$ & {MAE}$\da$ & {SSIM}$\ua$ & {PSNR}(dB)$\ua$ & {LPIPS}$\da$ && {FID}$\da$ & {FVD$_{16}$}$\da$ & {FVD$_{128}$}$\da$ & {IS}$\ua$ \\ 
        \cmidrule{2-6}\cmidrule{8-11}
        \gls{dyn}                & $3e\text{-}3_{\pm{8e\text{-}4}}$ & $0.03_{\pm{5e\text{-}3}}$ & $0.78_{\pm{0.05}}$ & $24.9_{\pm{1.08}}$ & $0.08_{\pm{0.01}}$ && $16.9$ & $87.4$ & 69.2       & $2.33_{\pm{0.13}}$ \\
        \gls{ped}(\gls{a4c})    & $1e\text{-}3_{\pm{7e\text{-}4}}$ & $0.02_{\pm{6e\text{-}3}}$ & $0.86_{\pm{0.05}}$ & $28.9_{\pm{1.95}}$ & $0.07_{\pm{0.02}}$ &&  $6.9$ & $21.8$ & -  & $2.90_{\pm{0.09}}$ \\
        \gls{ped}(\gls{psax})   & $2e\text{-}3_{\pm{8e\text{-}4}}$ & $0.02_{\pm{6e\text{-}3}}$ & $0.85_{\pm{0.05}}$ & $28.5_{\pm{2.14}}$ & $0.08_{\pm{0.02}}$ &&  $8.1$ & $22.1$ & -  & $3.16_{\pm{0.13}}$    \\
    \end{tabular}
    }
\end{table}

The \gls{lidm}s are evaluated using generative metrics \gls{fid} and \gls{is}. We generate 100,000 samples with each \gls{lidm}, and apply our privacy filter to remove any sample that has a close match in the corresponding real training set.
Then, we compute \gls{fid} and \gls{is} on the unfiltered and filtered \gls{vae}-decoded latents.
\Cref{table:lidm} shows that our privacy filter maintains the overall quality of the generated images. The `Rejected Samples' shows that our \gls{lidm}s have limited memorization.

\begin{table}[t]\setlength{\tabcolsep}{8pt}
    \centering
    \caption{Performance of the \gls{lidm} before and after applying privacy filtering.}
    \label{table:lidm}
    \resizebox{0.9\columnwidth}{!}{%
    \begin{tabular}{lccccccc}
        & \multicolumn{2}{c}{\textbf{Before Filtering}} & &\multicolumn{2}{c}{\textbf{After Filtering}} & &\textbf{Rejected}\\
        & {FID}$\da$ & {IS}$_{\pm\mathbf{std}}$$\ua$ && {FID}$\da$ & {IS}$_{\pm\mathbf{std}}$$\ua$ &&  \textbf{Samples}\\ 
        \cmidrule{2-3}\cmidrule{5-6}\cmidrule{8-8}
        \gls{dyn}               & 17.3  & $2.42_{\pm{0.02}}$   & & 16.4  & $2.37_{\pm{0.02}}$ && 11.25\%\\ %ddim 380k 64 100k samples
        \gls{ped}(\gls{a4c})    & 13.7  & $2.86_{\pm{0.03}}$    && 11.0  & $2.95_{\pm{0.03}}$ && 37.06\%\\ %ddpm 470k 64 100k samples
        \gls{ped}(\gls{psax})   & 16.8  & $3.05_{\pm{0.03}}$   & & 14.5  & $3.03_{\pm{0.02}}$ && 27.45\%\\ %ddpm 420k 64 100k samples
    \end{tabular}
    }
\end{table}

The privacy filtering models all reach Recall scores above 99\%, ensuring a low false-negative rate.
Combined with our distance thresholding method, they drastically reduce the risk of private data being present in the remaining samples.

\begin{figure}[t]
    \centering
    \includegraphics[width=\linewidth,trim={0cm 0 0cm 0},clip]{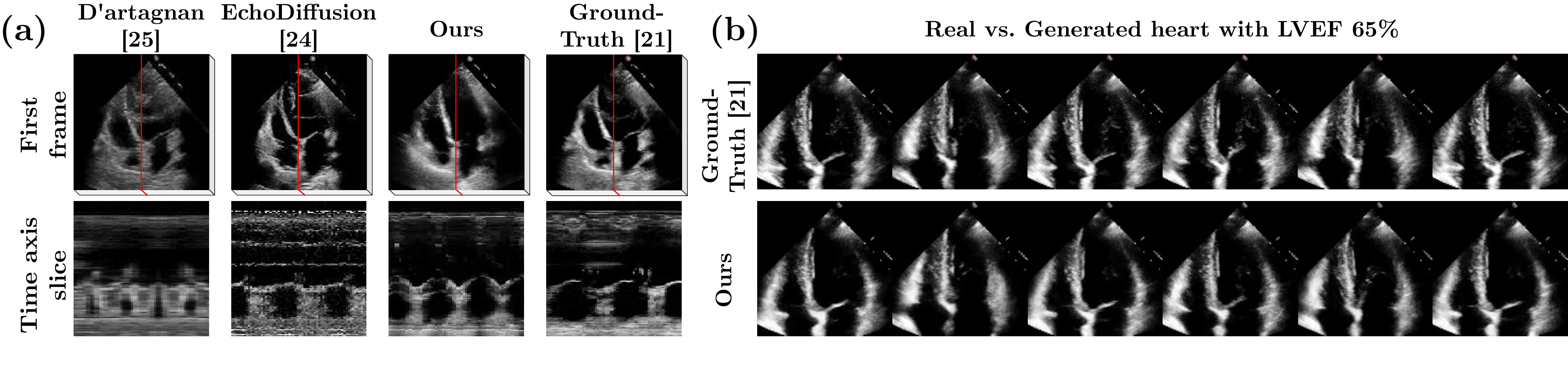}
    \caption{
    (a) Temporal consistency of our generated videos against previous works \cite{reynaud2023feature,reynaud2022d} and the real data~\cite{ouyang2020video}. (b) Qualitative results of our generated videos against the real data~\cite{ouyang2020video}, animating the same heart with the same \gls{ef}.
    }
    \label{fig:qualitative}
\end{figure}

To evaluate the \gls{lvdm}, we use generative metrics \gls{fid}, \gls{fvd}$_{16}$, \gls{fvd}$_{128}$ and \gls{is}. 
We generate 2048 videos with 192 frames for \gls{dyn} and 128 frames for \gls{ped}. 
We use two types of latent images to condition the video generation: (1) encoded real images, and (2) \gls{lidm}-generated latent images.
For the \gls{dyn} dataset, \Cref{table:lvdm} shows that our \gls{lvdm} performs better when using encoded real images than when using \gls{lidm}-generated latent images. 
Nevertheless, our \gls{lvdm} using \gls{lidm}-generated latent images achieves competitive results compared to the state-of-the-art echocardiogram generation method~\cite{reynaud2023feature}, in terms of image quality (\gls{fid}) and video quality (\gls{fvd}). 
Qualitatively, \Cref{fig:qualitative} (b) shows that our \gls{lvdm} can generate videos which are indistinguishable from real ones.
For the \gls{ped} dataset, \Cref{table:lvdm} shows that our \gls{lvdm} achieves good overall performance. 
Regarding video generation sampling time, our method (2.4s for 64 frames) is two orders of magnitude faster than the state-of-the-art method~\cite{reynaud2023feature} (279s for 64 frames). 
In addition, with the proposed video stitching strategy, we can generate infinitely long temporally-consistent videos. 
To demonstrate the scalability of our method, we generate a 10-minute-video (19,200 frames at 32 fps) in 14 minutes on a A100 GPU.

\begin{table}[t]
    \centering
    \caption{Performance of the \gls{lvdm} when generating videos from encoded real hearts (left) and \gls{lidm}-generated hearts (right). Sampling time for 64 frames, or 2s of video.
    }
    \label{table:lvdm}
    \resizebox{\columnwidth}{!}{%
    \begin{tabular}{lccccccccccc}
        & \multicolumn{4}{c}{\textbf{Real Hearts}} && \multicolumn{4}{c}{\textbf{LIDM Generated Hearts}} && \textbf{Sampling}\\
        & {FID}$\da$ &  {FVD}$_{16}\da$ & {FVD}$_{128}\da$ & {IS}$_{\pm\mathbf{std}}$$\ua$ && {FID}$\da$ & {FVD}$_{16}\da$ & {FVD}$_{128}\da$ & {IS}$_{\pm\mathbf{std}}$$\ua$ && \textbf{Time}\\ 
        \cmidrule{2-5}\cmidrule{7-10} \cmidrule{12-12}
        EchoDiff.\cite{reynaud2023feature} & 24.0  & 228 & -    & $2.59_{\pm{0.06}}$ && 28.0   & 294.4   & -   & $2.47_{\pm{0.04}}$  && 279s\\
        \gls{dyn}               & 17.4  & 71.4 & 168.3  & $2.31_{\pm{0.08}}$ && 28.8  & 103.5 & 280.2 & $2.26_{\pm{0.08}}$ && 2.4s\\
        \cmidrule{2-5}\cmidrule{7-10}\cmidrule{12-12}
        \gls{ped}(\gls{a4c})    & 24.8  & 112.2& -    & $2.69_{\pm{0.18}}$ && 22.6  & 86.6  & -   & $2.77_{\pm{0.23}}$ && 2.4s\\
        \gls{ped}(\gls{psax})   & 33.0  & 126.9& -    & $2.49_{\pm{0.09}}$ && 27.9  & 85.1  & -   & $2.86_{\pm{0.18}}$ && 2.4s\\
        %\hline
    \end{tabular}
    }
\end{table}

\noindent\textbf{EchoNet-Synthetic: }
We generate \gls{syn} with our video generation pipeline. \gls{syn} contains three sub-datasets, corresponding to the three real datasets. 
Each sub-dataset has the same number of videos as its corresponding real dataset, and the number of frames in each video is the average number of frame in the corresponding real dataset.
The visual quality of \gls{syn} is competitive with previous state-of-the-art, and is indistinguishable from real data (\Cref{fig:qualitative} (a) top row).
In terms of temporal consistency, \gls{syn} outperforms all previous works, and matches the real data (\Cref{fig:qualitative} (a) bottom row).

\noindent\textbf{Downstream Evaluation: }
We train \gls{ef} regression models on the \gls{syn} datasets. 
We use \gls{ef} prediction metrics (R2, MAE, RMSE) from \cite{ouyang2020video,reddy2023video} as baselines, where the models have been trained on the real data. 
We report the results from these works (`Claimed' in Table~\ref{table:downstream}) and reproduce them exactly with the same models and training setup (`Reproduced' in Table~\ref{table:downstream}).
To test the information loss caused by \gls{vae} encoding/decoding, we train the regression models on the \gls{vae}-reconstructed real data (`\gls{vae} rec.' in Table~\ref{table:downstream}). 
We also train the regression model on synthetic data generated with EchoDiffusion~\cite{reynaud2023feature}, the most recent ultrasound synthesis diffusion model (`EchoDiff.~\cite{reynaud2023feature}' in Table~\ref{table:downstream}).
The results on our \gls{syn} datasets are shown in row `Ours' in Table~\ref{table:downstream}.
From Table~\ref{table:downstream}, we observe that the regression models trained on \gls{syn} perform notably better than the state-of-the-art (\emph{i.e.}, `EchoDiff.~\cite{reynaud2023feature}' vs. `Ours'). 
The models trained on \gls{syn} overall perform well on the real data, although they are not equivalent to the models trained directly on the real data. 
We attribute this performance difference to residual domain shift, given the high performance of the regression model trained and tested on the \gls{syn} dataset (`Ours (\gls{syn})' in Table~\ref{table:downstream}).

\begin{table}[t]
    \centering
    \caption{Comparison for \gls{ef} regression models performance. 
    `Claimed' and `Reproduced' are the claimed results in~\cite{ouyang2020video,reddy2023video} and our reproduced results, respectively. 
    `\gls{vae} rec.', `EchoDiff.~\cite{reynaud2023feature}' and `Ours' are the results where the regression model is trained on \gls{vae}-reconstructed data, generated data from \cite{reynaud2023feature} and our generated data, respectively, and evaluated on the real data.
    The last row shows the regression model trained and tested on our \gls{syn} datasets.
    }
    \label{table:downstream}
    \resizebox{0.8\columnwidth}{!}{%
    \begin{tabular}{lccccccccccc}
        & \multicolumn{3}{c}{\textbf{Dynamic}~\cite{ouyang2020video}} & &\multicolumn{3}{c}{\textbf{Ped (A4C)}~\cite{reddy2023video}} && \multicolumn{3}{c}{\textbf{Ped (PSAX)}~\cite{reddy2023video}} \\
        & {R2}$\ua$ & {MAE}$\da$ & {RMSE}$\da$ && {R2}$\ua$ & {MAE}$\da$ & {RMSE}$\da$ && {R2}$\ua$ & {MAE}$\da$ & {RMSE}$\da$ \\
        \cmidrule{2-4}\cmidrule{6-8}\cmidrule{10-12}
        Claimed             & 0.81 & 4.05 & 5.32 && 0.70 & 4.15 & 5.70 && 0.74 & 3.80 & 5.14 \\
        Reproduced          & 0.81 & 3.98 & 5.29 && 0.68 & 4.19 & 5.71 && 0.71 & 3.79 & 5.14 \\
        \gls{vae} rec.      & 0.80 & 4.12 & 5.48 && 0.68 & 4.21 & 5.71 && 0.72 & 3.82 & 5.14 \\
        \cmidrule{2-4}\cmidrule{6-8}\cmidrule{10-12}
        EchoDiff.~\cite{reynaud2023feature} & 0.55 & 6.02 & 8.21 & & - & - & - & & - & - & - \\
        Ours       & 0.75 & 4.55 & 6.10    && 0.70 & 5.06  & 7.07  && 0.68 & 4.82  & 6.95 \\
        \midrule
        & \multicolumn{3}{c}{\textbf{Syn Dynamic}} & &\multicolumn{3}{c}{\textbf{Syn Ped (A4C)}} && \multicolumn{3}{c}{\textbf{Syn Ped (PSAX)}} \\
        \cmidrule{2-4}\cmidrule{6-8}\cmidrule{10-12}
        Ours (\gls{syn}) & 0.93 & 1.67 & 2.24       && 0.94 & 1.24  & 1.90  && 0.96 & 0.94  & 1.51 \\
    \end{tabular}
    }
\end{table}

\section{Conclusion}
In this paper, we introduced a new end-to-end protocol for surrogate medical dataset generation, showcasing its adaptability through the provision of open-source code and a synthetic dataset that facilitates replication and further investigation by the research community.
Our work not only highlights a novel approach to generating extended video sequences beyond the initial training scope of the video diffusion model but also integrates a cutting-edge privacy preservation technique to ensure the synthetic datasets can be shared safely.
We believe this to be a step forward in leveraging synthetic datasets, offering a method that maintains dataset attributes in terms of size, quality and diversity.
While we acknowledge the complexity surrounding intellectual property, it is primarily a political challenge, and our focus here remains on the technical advancements that should mitigate the legal constraints around patient data safety. 
We are committed to employing the here presented protocol to release synthetic surrogates of private datasets in the future.

\noindent\textbf{Acknowledgements:} 
This work was supported by Ultromics Ltd., the UKRI Centre for Doctoral Training in Artificial Intelligence for Healthcare  (EP / S023283/1) and HPC resources provided by the Erlangen National High Performance Computing Center (NHR@FAU) of the Friedrich-Alexander-Universität Erlangen-Nürnberg (FAU) under the NHR project b180dc. NHR and high-tech agenda Bavaria (HTA) funding is partly provided by federal and Bavarian state authorities. NHR@FAU hardware is partially funded by the German Research Foundation (DFG) – 440719683. Support was also received from the ERC - project MIA-NORMAL 101083647 and DFG KA 5801/2-1, INST 90/1351-1.

\bibliographystyle{splncs04}
\bibliography{bibliography.bib}

\end{document}